\documentclass[letterpaper]{article} 
\usepackage{aaai25}  
\usepackage{times}  
\usepackage{helvet}  
\usepackage{courier}  
\usepackage[hyphens]{url}  
\usepackage{graphicx} 
\def\UrlFont{\rm}  
\usepackage{natbib}  
\usepackage{caption} 
\usepackage{algorithm}
\usepackage{algorithmic}

\usepackage{enumitem}
\usepackage{xcolor}
\usepackage{array}
\usepackage{tabularx}
\usepackage{booktabs}
\usepackage{multirow}
\usepackage{amsmath}
\usepackage{amssymb}
\usepackage{amsfonts} 
\usepackage{subcaption}

\usepackage{newfloat}
\usepackage{listings}
\DeclareCaptionStyle{ruled}{labelfont=normalfont,labelsep=colon,strut=off} 
\floatstyle{ruled}
\newfloat{listing}{tb}{lst}{}
\floatname{listing}{Listing}
\usepackage{natbib}

\title{Guided Learning: Lubricating End-to-End Modeling for Multi-stage Decision-making}
\author{
    Jian Guo\textsuperscript{\rm 1}\equalcontrib\thanks{Corresponding author}, Saizhuo Wang\textsuperscript{\rm 2}\equalcontrib\thanks{Work done during internship at IDEA Research}, Yiyan Qi\textsuperscript{\rm 1}\equalcontrib
}

\affiliations{
    \textsuperscript{\rm 1}IDEA Research 
    \textsuperscript{\rm 2}The Hong Kong University of Science and Technology

    \{guojian, qiyiyan\}@idea.edu.cn, swangeh@connect.ust.hk
}

\begin{document}

\maketitle
\begin{abstract}
Multi-stage decision-making is crucial in various real-world artificial intelligence applications, including recommendation systems, autonomous driving, and quantitative investment systems. In quantitative investment, for example, the process typically involves several sequential stages such as factor mining, alpha prediction, portfolio optimization, and sometimes order execution. While state-of-the-art end-to-end modeling aims to unify these stages into a single global framework, it faces significant challenges: (1) training such a unified neural network consisting of multiple stages between initial inputs and final outputs often leads to suboptimal solutions, or even collapse, and (2) many decision-making scenarios are not easily reducible to standard prediction problems. To overcome these challenges, we propose Guided Learning, a novel methodological framework designed to enhance end-to-end learning in multi-stage decision-making. We introduce the concept of a ``guide'', a function that induces the training of intermediate neural network layers towards some phased goals, directing gradients away from suboptimal collapse. For decision scenarios lacking explicit supervisory labels, we incorporate a utility function that quantifies the ``reward'' of the throughout decision. Additionally, we explore the connections between Guided Learning and classic machine learning paradigms such as supervised, unsupervised, semi-supervised, multi-task, and reinforcement learning. Experiments on quantitative investment strategy building demonstrate that guided learning significantly outperforms both traditional stage-wise approaches and existing end-to-end methods.

\end{abstract}

\newcommand{\figureExample}{
    \begin{figure*}[ht]
    \centering
    \includegraphics[width=\textwidth]{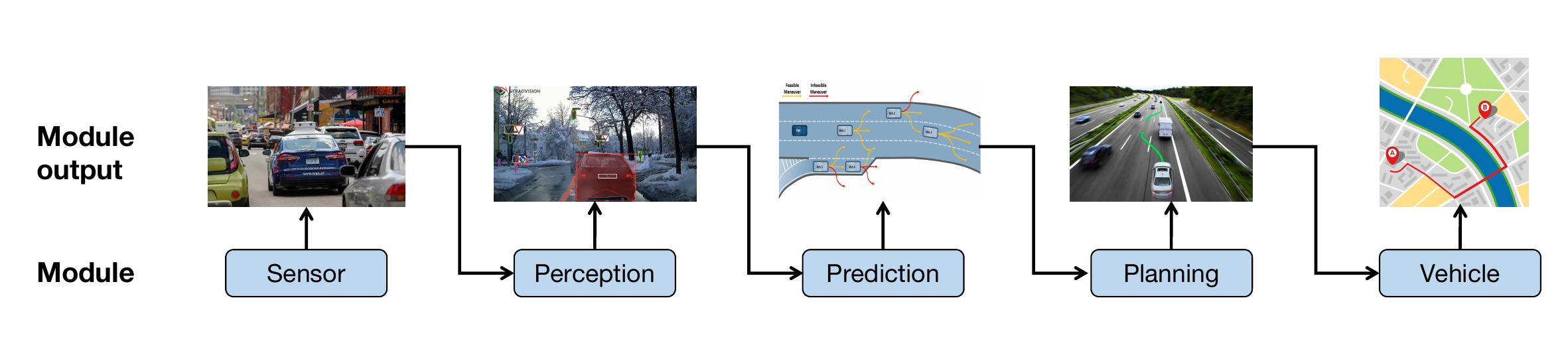}
    \caption{Example pipeline in autonomous driving systems}
    \label{fig:Example}
    \end{figure*}
}

\newcommand{\figureEquPipeline}{
    \begin{figure*}[ht]
    \centering
    \includegraphics[width=\textwidth]{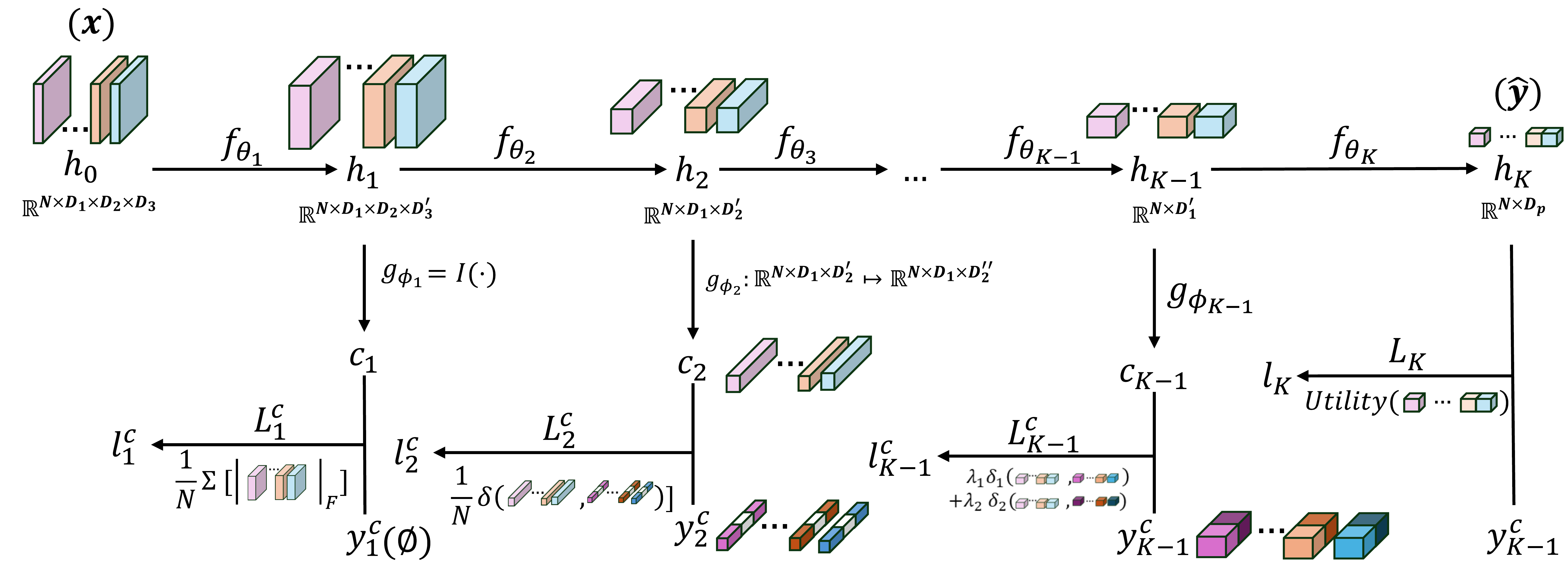}
    \caption{
    Illustration of guided learning in an end-to-end model with an example guide implementation. The model is conceptually separated into $K$ stages, each represented by a function $f_{\theta_k}$ that transforms $h_{k-1}$ to $h_k$. At each stage, a guided head $g_{\phi_k}$ maps $h_k$ to a phased output $c_k$. A guided loss function $L^c_k$ is computed at each stage with different implementations. For example, $L^c_1$ is an unsupervised loss based on feature norms. $L^c_2$ represents a loss defined on partial features using a distance metric $\delta(\cdot, \cdot)$. $L^c_{K-1}$ represents a weighted sum of multiple losses computed with labels in multiple perspectives (e.g. detection loss and segmentation loss), as illustrated in different colors. The pipeline produces a final output $\hat{y}$ and computes a utility-based loss $L_K$ for all samples.
    }
    \label{fig:EquPipeline}
    \end{figure*}
}

\newcommand{\figurePortfolioOptimization}{
\begin{figure}[!t]
    \centering
    \includegraphics[width=\linewidth]{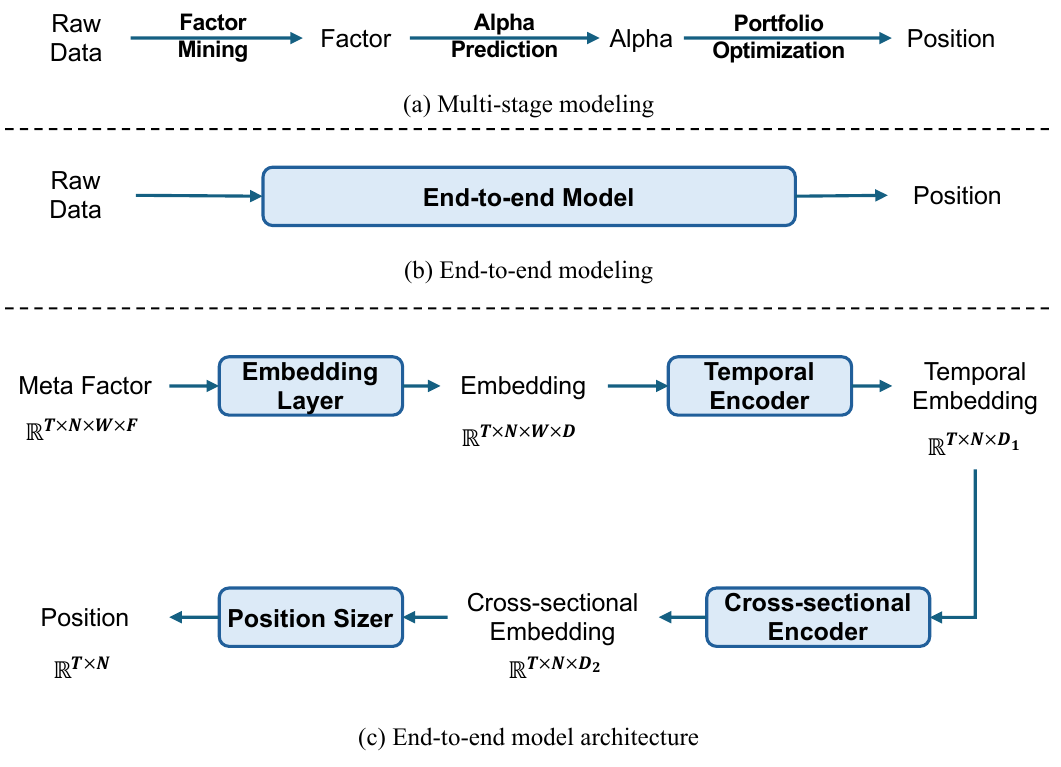}
    \caption{Illustration of the multi-stage approach and end-to-end approach for quantitative investment.}
    \label{fig:PortfolioBackground}
\end{figure}
}

\newcommand{\figureEndToEndPortfolioModelArch}{
\begin{figure}[ht]
    \centering
    \label{fig:End2EndPortfolioModelArch}
    \includegraphics[width=\linewidth]{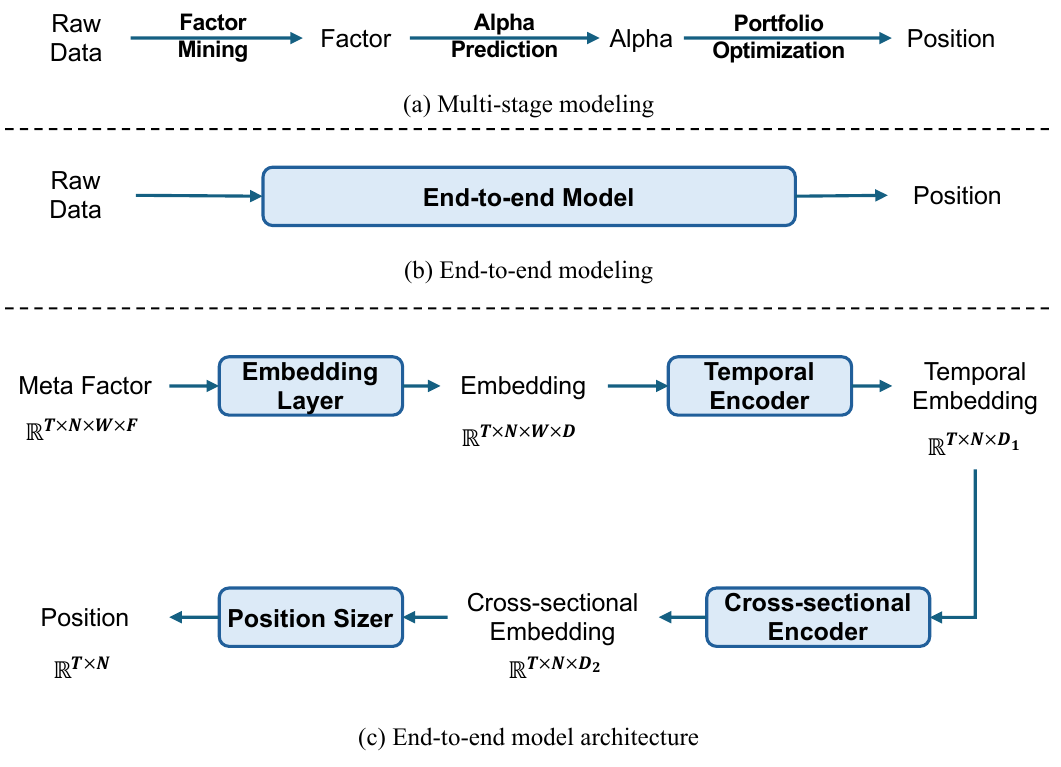}
    \caption{Illustration of the multi-stage approach and end-to-end approach for portfolio construction.}
\end{figure}
}

\newcommand{\figurePsen}{
\begin{figure*}[!h]
    \centering
    \begin{subfigure}{0.24\textwidth}
        \includegraphics[width=\linewidth]{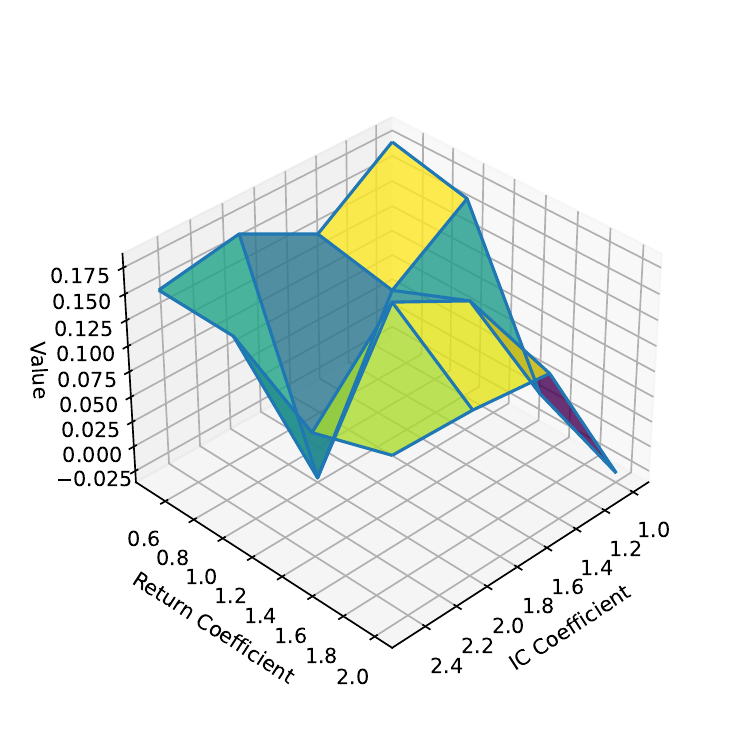}
        \caption{Annualized return}
    \end{subfigure}
    \begin{subfigure}{0.24\textwidth}
        \includegraphics[width=\linewidth]{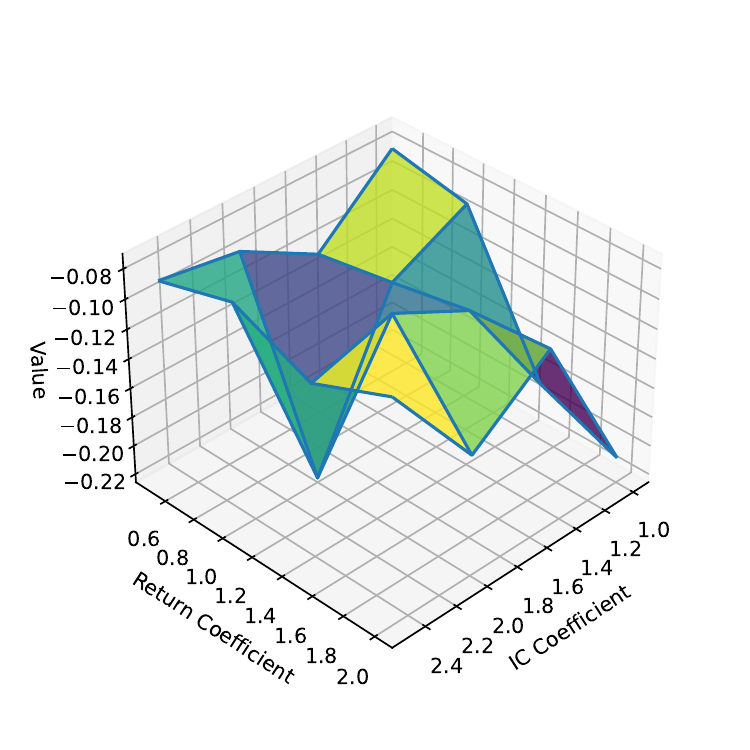}
        \caption{Maximum drawdown}
    \end{subfigure}
    \begin{subfigure}{0.24\textwidth}
        \includegraphics[width=\linewidth]{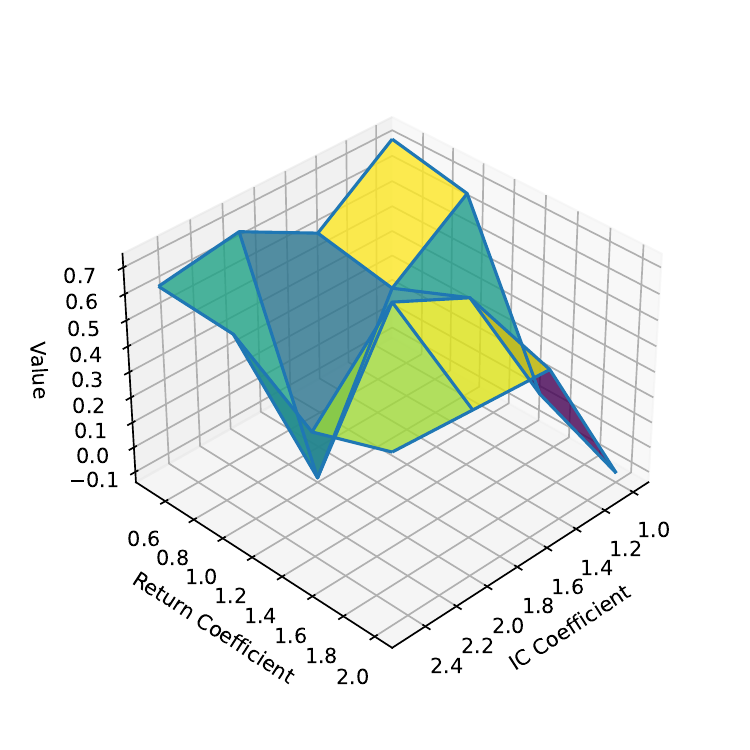}
        \caption{Sharpe ratio}
    \end{subfigure}
    \begin{subfigure}{0.24\textwidth}
        \includegraphics[width=\linewidth]{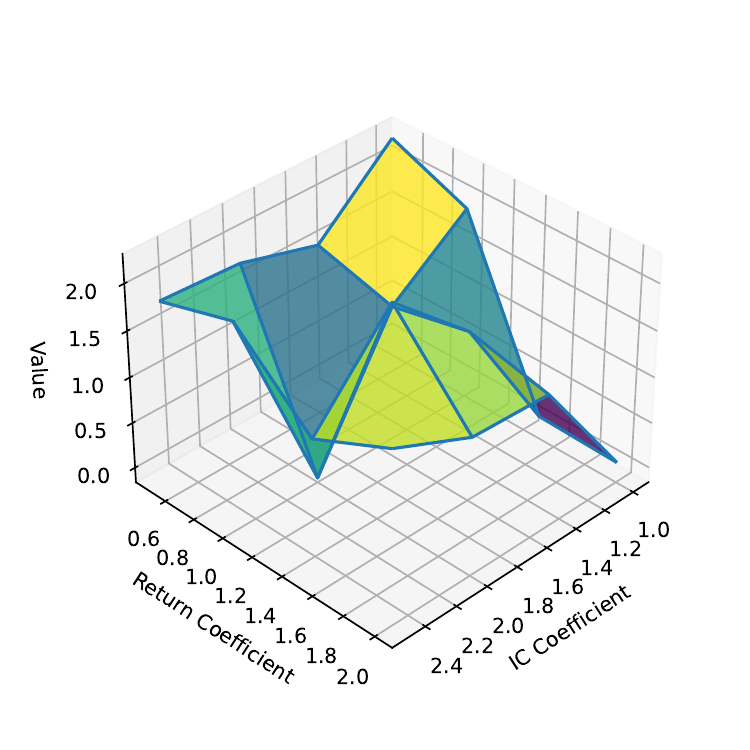}
        \caption{Calmar ratio}
    \end{subfigure}
    \caption{Parameter sensitivity}
    \label{fig:psen}
\end{figure*}
}

\newcommand{\figureBacktest}{
\begin{figure*}[ht]
    \centering
    \includegraphics[width=0.9\textwidth]{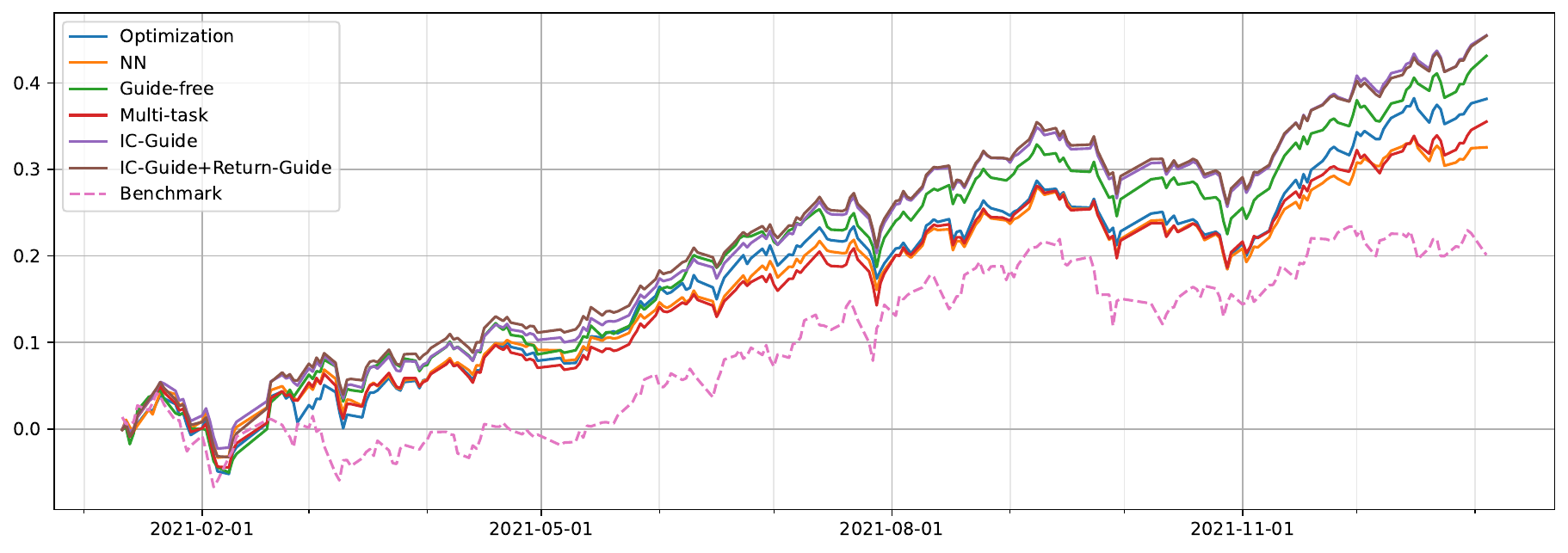}
    \caption{Backtest curve}
    \label{fig:backtest}
\end{figure*}
}

\newcommand{\tableOverallEffectiveness}{
    \begin{table*}[ht]
      \centering
      \caption{Overall effectiveness of guided learning}
        \begin{tabular}{cl|cccc}
        \toprule
              &       & \textbf{Ann. Return} &\textbf{ Max. Drawdown} & \textbf{Sharpe Ratio} & \textbf{Calmar Ratio} \\
        \midrule
        \multirow{2}[2]{*}{Stage-wise} & Optimization & 17.97\% $\pm$ 9.70\%  & -10.90\% $\pm$ 3.29\% & 0.73 $\pm$ 0.38  & 1.86 $\pm$ 0.92  \\
              & NN    & 12.41\% $\pm$ 5.98\% & -9.30\% $\pm$ 2.45\% & 0.52 $\pm$ 0.25   & 1.54 $\pm$ 1.05  \\
        \midrule
        \multirow{4}[2]{*}{Guided} & Guide-free  &   23.19\% $\pm$ 0.84\%  & -8.90\% $\pm$ 0.39\% & 0.94 $\pm$ 0.03  & 2.58 $\pm$ 0.12  \\
              & Multi-task &   15.33\% $\pm$ 1.55\%  & -8.20\% $\pm$ 0.53\% & 0.62 $\pm$ 0.09  & 1.63 $\pm$0.72  \\
              & IC-Guide &   \textbf{25.28\%} $\pm$ 2.10\% & \textbf{-7.70\%} $\pm$ 0.81\% & 1.05 $\pm$ 0.09   & \textbf{3.27} $\pm$ 0.25  \\
              & IC-Guide + Return Guide &   25.24\% $\pm$ 3.22\%  & -7.80\% $\pm$ 0.50\% & \textbf{1.06} $\pm$ 0.13 & 3.23 $\pm$ 0.22  \\
        \bottomrule
        \end{tabular}%
    \label{tab:overall_effectiveness}%
    \end{table*}%
}

\newcommand{\tableModelGeneralize}{
\begin{table*}[ht]
  \centering
  \caption{Performance comparison across different temporal encoder architectures}
    \begin{tabular}{cc|cccc}
    \toprule
          &       & \textbf{Ann. Return} & \textbf{Max. Drawdown} & \textbf{Sharpe Ratio} & \textbf{Calmar Ratio} \\
    \midrule
    \multirow{3}[2]{*}{LSTM} & -     & 13.00\% $\pm$ 7.92\% & -10.06\% $\pm$ 1.98\% & 0.54 $\pm$ 0.33 & 1.41 $\pm$ 0.91 \\
          & + IC Guide & 13.68\% $\pm$ 2.66\% & -9.37\% $\pm$ 3.10\% & 0.56 $\pm$ 0.11 & 1.58 $\pm$ 0.55 \\
          & \textcolor[rgb]{ .235,  .49,  .133}{Improve} & \textcolor[rgb]{ .235,  .49,  .133}{5.23\%} & \textcolor[rgb]{ .235,  .49,  .133}{6.85\%} & \textcolor[rgb]{ .235,  .49,  .133}{3.59\%} & \textcolor[rgb]{ .235,  .49,  .133}{12.30\%} \\
    \midrule
    \multirow{3}[2]{*}{TCN} & -     & 19.13\% $\pm$ 3.10\% & -8.63\% $\pm$ 0.72\% & 0.79 $\pm$ 0.13 & 2.21 $\pm$ 0.19 \\
          & + IC Guide & 20.57\% $\pm$ 5.53\% & -8.77\% $\pm$ 1.53\% & 0.84 $\pm$ 0.23 & 2.47 $\pm$ 0.91 \\
          & \textcolor[rgb]{ .235,  .49,  .133}{Improve} & \textcolor[rgb]{ .235,  .49,  .133}{7.55\%} & \textcolor[rgb]{ .235,  .49,  .133}{-1.61\%} & \textcolor[rgb]{ .235,  .49,  .133}{6.59\%} & \textcolor[rgb]{ .235,  .49,  .133}{11.84\%} \\
    \midrule
    \multirow{3}[2]{*}{PatchTST} & -     & 7.73\% $\pm$ 0.10\% & -10.92\% $\pm$ 0.04\% & 0.34 $\pm$ 0.00 & 0.71 $\pm$ 0.01 \\
          & + IC Guide & 7.95\% $\pm$ 2.32\% & -11.40\% $\pm$ 1.32\% & 0.34 $\pm$ 0.10 & 0.72 $\pm$ 0.24 \\
          & \textcolor[rgb]{ .235,  .49,  .133}{Improve} & \textcolor[rgb]{ .235,  .49,  .133}{2.85\%} & \textcolor[rgb]{ .235,  .49,  .133}{-4.33\%} & \textcolor[rgb]{ .235,  .49,  .133}{2.39\%} & \textcolor[rgb]{ .235,  .49,  .133}{1.75\%} \\
    \bottomrule
    \end{tabular}%
  \label{tab:model_generalize}%
\end{table*}%
}

\newcommand{\tableWhereToAdd}{
\begin{table*}[ht]
  \centering
  \caption{Performance comparison across different guide positions}
    \begin{tabular}{l|cccc}
    \toprule
          & \multicolumn{1}{c}{\textbf{Ann. Return}} & \multicolumn{1}{c}{\textbf{Max. Drawdown}} & \multicolumn{1}{c}{\textbf{Sharpe Ratio}} & \multicolumn{1}{c}{\textbf{Calmar Ratio}} \\
    \midrule
    Embedding & 20.75\% $\pm$ 3.76\% & -8.38\% $\pm$ 1.25\% & 0.86 $\pm$ 0.16 & 2.51 $\pm$ 0.49 \\
    Temporal & 21.18\% $\pm$ 2.19\% & -7.77\% $\pm$ 0.81\% & 0.88 $\pm$ 0.09 & 2.73 $\pm$ 0.17 \\
    Cross-sectional   & 19.29\% $\pm$ 7.90\% & -9.35\% $\pm$ 1.65\% & 0.80 $\pm$ 0.34 & 2.19 $\pm$ 1.02 \\
    \bottomrule
    \end{tabular}%
  \label{tab:where_to_add}%
\end{table*}%
}

\newcommand{\tableWhatToAdd}{
\begin{table*}[ht]
  \centering
  \caption{Performance comparison across different types of guide}
    \begin{tabular}{c|cccc}
    \toprule
          & \multicolumn{1}{c}{\textbf{Ann. Return}} & \multicolumn{1}{c}{\textbf{Max. Drawdown}} & \multicolumn{1}{c}{\textbf{Sharpe Ratio}} & \multicolumn{1}{c}{\textbf{Calmar Ratio}} \\
    \midrule
    MSE   & 18.94\% $\pm$ 2.95\% & -9.21\% $\pm$ 0.29\% & 0.78 $\pm$ 0.12 & 2.06 $\pm$ 0.37 \\
    CLF   & 17.75\% $\pm$ 7.41\% & -8.89\% $\pm$ 0.39\% & 0.73 $\pm$ 0.30 & 2.00 $\pm$ 0.84 \\
    Rank  & 7.66\% $\pm$ 0.00\% & -10.95\% $\pm$ 0.00\% & 0.33 $\pm$ 0.00 & 0.70 $\pm$ 0.00 \\
    \bottomrule
    \end{tabular}%
  \label{tab:what_to_add}%
\end{table*}%
}

\newcommand{\tableCalibrationInteraction}{
\begin{table}[ht]
  \centering
  \caption{Interaction between calibration objectives}
    \begin{tabular}{lllrrrrr}
    \multicolumn{3}{c}{Setting} & \multicolumn{1}{c}{ARR} & \multicolumn{1}{c}{SP} & \multicolumn{1}{c}{TR} & \multicolumn{1}{c}{MDD} & \multicolumn{1}{c}{Var} \\
    Cali1 & Cali2 & Cali3 &       &       &       &       &  \\
    \checkmark &       &       &       &       &       &       &  \\
          & \checkmark &       &       &       &       &       &  \\
          &       & \checkmark &       &       &       &       &  \\
    \checkmark & \checkmark &       &       &       &       &       &  \\
          & \checkmark & \checkmark &       &       &       &       &  \\
    \checkmark &       & \checkmark &       &       &       &       &  \\
    \checkmark & \checkmark & \checkmark &       &       &       &       &  \\
    \end{tabular}%
  \label{tab:addlabel}%
\end{table}%
}
\section{Introduction}
Many real-world AI systems depend on multi-stage decision-making, where a sequence of predictions/decisions is made in order to output the final decision. Examples include search engine \cite{almukhtar2021search}, recommendation systems \cite{ko2022survey}, dialog system \cite{mctear2022conversational}, robotics \cite{zhao2020sim}, autonomous driving \cite{chen2024end}, drones \cite{kangunde2021review}, AIGC systems \cite{foo2023ai}, and quantitative trading \cite{wei_e2eai_2023}.

\figureExample

\begin{itemize}[leftmargin=*]
    \item \textbf{Example 1:} Traditional autonomous driving systems (Figure \ref{fig:Example}) consist of three key sequential stages: perception, prediction, and planning. The perception stage detects traffic lights and other critical elements, followed by the prediction stage, which forecasts the potential movements of surrounding entities. Finally, the planning module ensures that the vehicle's path prioritizes both safety and efficiency.
    
    \item \textbf{Example 2:} In quantitative finance, the research pipeline typically includes the following sequential stages: factor mining (extracting informative features or trading signals), alpha prediction (building machine learning models to predict alphas), and portfolio optimization (finding the optimal asset positions to balance portfolio return and risk).
\end{itemize}

Although traditional stage-wise approaches for multi-stage decision-making offer advantages such as transparency and flexibility, they suffer from several limitations. For instance, the optimization goals across these stages often lack consistency, leading to sub-optimal final solutions. Moreover, errors in one stage can accumulate and propagate through subsequent stages. 

As an alternative to stage-wise approaches, end-to-end deep learning has demonstrated its efficiency and effectiveness in modeling by processing initial inputs directly to produce final outputs. However, end-to-end modeling faces significant challenges as well. Firstly, neural networks designed for end-to-end learning in multi-stage decision-making problems are typically deep, making model training and parameter tuning difficult. Secondly, many scenarios cannot be easily formulated as standard supervised learning problems. For example, when applying end-to-end modeling to a quantitative investment pipeline, it is challenging to explicitly formulate it as a supervised learning problem because ground-truth optimal positions are not observable and positions depend on the alphas predicted in previous layers of the end-to-end neural network.

To address these issues, we propose Guided Learning (GL), a new methodological framework for smoothing the end-to-end learning for multi-stage decision-making. 

Our main contributions are summarized as:
\begin{itemize}[leftmargin=*]
\item We propose a general machine learning framework that enhances the training effectiveness of end-to-end learning for multi-stage decision-making.
\item We introduce a new machine learning concept called ``guide'', a function that supervises the training of intermediate neural network layers, directing gradients away from suboptimal collapse. 
\item For decision scenarios lacking explicit supervisory labels, we incorporate a utility function that quantifies the ``reward'' of the throughout decision. 
\item Additionally, we explore the connections between Guided Learning and classic machine learning paradigms such as supervised, unsupervised, semi-supervised, multi-task, and reinforcement learning.
\end{itemize}

\section{Guided Learning}
In this section, we first formally define the learning paradigms of stage-wise learning and end-to-end learning. Next, we provide detailed training procedures for our guided learning approach. Finally, we discuss the relationship between our guided learning framework and other existing learning frameworks.

\subsection{Preliminary}
Many complex systems in real-world applications can be modeled as multi-stage decision problems.
The \textbf{stage-wise learning} paradigm handles each stage by a separate model. These models are trained independently with unique objectives, forming a pipeline where the output of one model serves as the input for the next. This process can be formalized as:
$$
x^{(i)} = q_i(x^{(i-1)}; \theta_i), \quad i = 1, \ldots, L
$$
Here, the pipeline is divided into $L$ stages. The terms $x^{(i-1)}$ and $x^{(i)}$ represent the input and output of the $i$-th stage's model, respectively. 
The overall dataset is denoted as $\mathcal{D}=\{\mathcal{D}_1, \mathcal{D}_2,\ldots, \mathcal{D}_L\}$, where $\mathcal{D}_1$ is the train dataset of the $i$-th stage model. Each stage's model, parameterized by $\theta_i$, is trained separately as follows:
$$
\theta_1^* = \arg\min_{\theta_1} \mathcal{L}_1(\theta_1, \mathcal{D}_1)
$$
$$
\theta_2^* = \arg\min_{\theta_2} \mathcal{L}_2(\theta_2, \mathcal{D}_2)
$$
$$\cdots$$
The overall goal of the entire pipeline is then formulated through the following utility function:
\begin{equation}
    \theta_L^* = \arg\max_{\theta_L} U(\mathcal{D}_L; \theta_L),
\end{equation}
where $\theta_L^*$ is the optimal parameters of the $L$-th stage's model, $U$ is the utility function, often a prediction error or cost to be minimized. 
For traditional supervised learning, $U$ is the expected utility over all samples.
Although this pipelined approach can produce satisfactory outcomes, it has an inherent limitation. The training objectives of the intermediate models ($\mathcal{L}_i$) might not align perfectly with the final optimization objective $U(\cdot; \cdot)$. This mismatch is referred to as the prediction-optimization gap \cite{yan_surrogate_2021}.

\textbf{End-to-end learning} has been proposed to align the training objective with the final goal.
In an end-to-end learning paradigm, the whole problem is handled by a single model, parameterized by $\theta$. The training objective can be formulated as:
\begin{equation}
    \theta^* = \arg\max_{\theta} U(\mathcal{D}, \theta).
\end{equation}
However, directly optimizing a unified model can be unstable due to the complexity of integrating multiple stages from the original input to the final outputs. 

\figureEquPipeline

\subsection{Formulation of Guided Learning}
Considering the two approaches discussed above, our guided learning separates the end-to-end model into $L$ conceptual stages. Each stage is considered as handled by part of the machine learning model's parameter. In this way, the whole model can be separated into $L$ phases. 
Let $h_i \in \mathbb{R}^{N \times d_i}$ denote the output of the $i$-th stage, where $N$ is the sample size and $d_i$ is the feature dimension of the $i$-th stage. Specifically, $h_0=x$ is the raw input, and $h_N=\hat{y}$ is the final output of the model. By definition, we have:
\begin{equation}
    h_i = f_{\theta_i} (h_{i-1}), \quad i = 1,...,L
\end{equation}
The training objective is to maximize the final goal $U(h_N)$. Guided learning uses intermediate 'guidance' at the middle stages to prevent intermediate representations from collapsing while maintaining the flexibility of end-to-end learning.



\subsubsection{Guided Objective}
The guided objective at an intermediate stage $i$ consists of the following components: an optional guided head $g_{\phi_i}$, the phased output $c_i$, the phased goal $y^c_i$, and the guided loss function $l_i^c(\cdot, \cdot)$ on $c_i$ and $y^c_i$. In the following, we will describe each component in detail.

\paragraph{Phased Output}
The phased output is optionally computed by a guided head with additional parameters. This is for the flexibility of the guidance signal. For example, when $h_i$ is a high-dimensional embedding while we have scalar phased goals, a guided head is needed to extract relevant information from this latent embedding. Meanwhile, the guided head $g_{\phi_i}$ can characterize an identity transformation, which does not cause changes in the latent embedding. Having non-identity mapping may introduce new parameters into the original end-to-end model, and designing this guided head to minimize the additional overhead while achieving the phased goal is crucial.

\paragraph{Phased Goal}
The phased goal $y_i^c$ can be defined in multiple manners: for each sample, for each dimension, or for both each sample and each dimension. For example, a scalar phased goal can be set as values on the first dimension and zeros on other dimensions. In this way, each sample has a scalar phased goal. Meanwhile, some high-dimensional phased goal can also be represented in this form. For example, in an end-to-end vision model with object-detection guide, the high-dimensional phased goal represents the coordinates and size of bounding boxes. Moreover, additional masks on each sample/dimension can also be applied by setting part of the phased goal to zero at some positions. 

\paragraph{Guided Loss}
The guided loss is defined as very flexible and can take an arbitrary form. It can incorporate supervised learning, unsupervised learning, semi-supervised learning, etc.

The whole model is trained to minimize a composite loss function comprising both guided objective and the final objective:
\begin{equation}
    L = \sum_{i=1}^{K-1} \lambda_i L_i^c (c_i, y_i^c) + U(h_K)
\end{equation}
where $\lambda_i$ is the weighting coefficient for each guided loss term.

\subsection{Discussion}
In this section, we first discuss the impact of our proposed phased goal on the final objective. Then, we compare our guided learning with other learning paradigms with similar architecture.

\subsubsection{Influence of Guide}
In end-to-end learning, particularly in complex tasks such as multi-step investment strategy building, the loss landscape can be irregular. Adding phased goals aims to smooth this landscape, potentially guiding optimization to better parameter spaces and enhancing performance. However, a poor guide might lead to suboptimal areas, reduce performance, or have no significant impact if it adds no new information. The outcome depends on the specific problem, model architecture, and guidance choices, necessitating empirical studies to predict and manage these effects.
In our experiments, we investigate multiple guidance choices in the domain of portfolio optimization and then give some insight into how to select a better guide.

\subsubsection{Compared with Other Learning Paradigm}
In this section, we compare our guided learning approach with existing learning paradigms that have similar architectures.
\begin{itemize}[leftmargin=*]
    \item \textbf{Multi-task learning.} Multi-task learning (MTL) \cite{zhang2021survey} and guided Learning both aim to enhance model performance by leveraging multiple objectives. However, several distinctions set them apart:
1) Sub-objective placement: MTL typically confines all auxiliary tasks to the final layer, using them to restrict the search space for a shared feature encoder. Guided Learning, on the other hand, distributes its objectives throughout the network according to the stages.
2) Parameter Sharing: MTL often involves extensive parameter sharing across tasks, which can result in a significant seesaw phenomenon \cite{yang9personalized}, where performance improvements in one task lead to declines in another. Guided Learning, on the other hand, adopts a more flexible approach that allows for stage-specific guidance, thereby diminishing such conflicts.
Our experiments demonstrate that Guided Learning can more efficiently enhance the main task's performance compared to MTL.
    
    \item \textbf{Deep supervision.} Deep Supervision \cite{wang2015training} and guided learning both utilize intermediate loss to regularize the learning process. However, as illustrated in Figure \ref{fig:EquPipeline}, deep supervision regularizes each sample independently. In contrast, our guided learning approach employs diverse, stage-specific guidance objectives. These guidance objectives can target subsets of input samples, allowing for more tailored and effective regulation at different stages of the network.

    \item \textbf{Reinforcement learning.}
    Guided learning and reinforcement learning (RL) differ in their approach to multi-step decision problems. While RL assigns credit to individual steps and optimizes them separately, guided learning maintains a holistic view of the trajectory, optimizing all steps simultaneously while using guides to enhance model training. This distinction has several implications: guided learning may better preserve temporal coherence in long-term strategies, potentially achieve higher sample efficiency (especially when the final outcome is more informative than intermediate rewards), and offer more stable learning in problems with sparse or delayed rewards. Additionally, the use of guides provides flexibility in incorporating domain knowledge, which can be challenging to integrate into RL via reward shaping. 
\end{itemize}


\figurePortfolioOptimization

\section{Experiment}
In this section, we present an empirical study of guided learning applied to a real-world quantitative investment problem. We will first introduce the background of the problem, followed by an analysis of the experimental results to evaluate the effectiveness of guided learning.

\subsection{Problem Background}
We select for the experiment the problem of cross-sectional investment strategy building . The objective is to make sequential (usually in hundreds of steps) investment decisions from a universe of financial instruments, with the aim of maximizing portfolio performance as measured by risk-adjusted metrics.

As illustrated in Figure \ref{fig:PortfolioBackground}, the traditional approach to this problem involves multiple stages: factor mining to extract relevant features from raw financial data, alpha prediction to forecast returns, and portfolio construction to determine optimal asset weights. While widely used in practice, this multi-stage approach typically requires substantial human expertise to optimize each component independently. Furthermore, the separate objectives for each stage may lead to misalignment in the overall process.

In response to these limitations, end-to-end learning approaches that directly map raw data to portfolio positions have also been studied. This method reduces the need for manual feature engineering and aligns the entire process with the final objective. However, training such end-to-end models presents new challenges due to the extended pipeline it covers and the instability in directly optimizing risk-adjusted portfolio metrics, which involve the standard deviation of portfolio returns over multiple holding periods.
\tableOverallEffectiveness
\figureBacktest

\paragraph{Formulation}
Consider features of a cross-section as a sliding window of stock data, $x \in \mathbb{R}^{N\times W\times F}$, where $N$ is the number of instruments in a cross-section, $W$ is the size of the lookback window, and $F$ is the number of features (e.g. open price, volume, etc.). In multi-stage methods, a predictive model $f_\theta$ parameterized by $\theta$ takes as input $x$ and predicts the cross-sectional expected return  $\hat{y} \in \mathbb{R}^N$. This prediction is then used in a portfolio optimizer to generate the final positions $w \in [0, 1]^N$ by solving a convex quadratic optimization problem \cite{markowitz_portfolio_1952}. The model is trained by minimizing the prediction error, formulated as
\begin{equation}
    \theta^* = \arg\min_{\theta} \mathbb{E}_{x \sim D} [\delta (f_\theta(x), y) ]
\end{equation}
where $D$ is the training set consisting of historical cross-sections, $\delta(\cdot, \cdot)$ is certain distance metric such as mean-squared error or negative Pearson correlation coefficient (also known as information coefficient, IC), and $y$ is the label that is usually set as the actual future return. However, this training objective does not correspond to the final goal, risk-adjusted performance, such as the Sharpe ratio \cite{sharpe_mutual_1966}.
Therefore, in end-to-end modeling, the model $g_\phi$ parametrized by $\phi$ is expected to take as input the whole history used for training $X \in \mathbb{R}^{T\times N \times W \times F}$ and directly generates consecutive positions $W \in [0, 1]^{T \times N}$, and is trained by maximizing the performance across whole history
\begin{equation}
    \label{equ:orig_e2e_obj}
    \phi^* = \arg\max_{\phi} U(W, R)
\end{equation}
where $U$ is a utility function such as Sharpe ratio, and $R \in \mathbb{R}^{T\times N}$ is the actual return. In practice, due to computation complexity and memory limit, the objective in Eq. (\ref{equ:orig_e2e_obj}) cannot be directly optimized. It is then approximated by subsampling on history to get $x \in \mathbb{R}^{T' \times N \times W \times D}$, where $T'$ is smaller than the length of the whole training set but also long enough for carrying out a reasonable estimate of the risk-adjusted return for stochastic optimization.

To implement guided learning in our end-to-end model, we conceptually divide the model into multiple interconnected components, as illustrated in Figure \ref{fig:PortfolioBackground}(c). This modular approach allows us to apply guides to intermediate representations at various stages of the model. For example, we can apply guides to the temporal embedding, enhancing its ability to predict absolute trends of individual financial instruments. Similarly, guides can be applied to the cross-sectional embedding, improving its capacity to maximize returns for each individual cross-section.
By introducing these guides, we aim to leverage domain-specific knowledge and improve the model's performance at different stages of the investment process while maintaining the end-to-end nature.

\tableModelGeneralize
\tableWhereToAdd

\subsection{Experimental Setup}

\paragraph{Dataset}
We conducted our experiments using historical data from the Chinese A-share stock market, spanning from 2013-01-01 to 2022-01-04. The dataset was divided as follows: training set (2013-01-01 to 2020-06-30), validation set (2020-07-01 to 2020-12-31), and test set (the entire year of 2021). We utilized 500 proprietary meta features as initial inputs, including basic volume-price data and fundamental indicators. Prior to model input, we performed necessary data preprocessing and filtering.

\paragraph{Model Architecture}

The end-to-end model architecture follows the design illustrated in Figure \ref{fig:PortfolioBackground}(c). For comparison, we also implemented a multi-stage predictive model consisting of an embedding layer, a temporal module, and a prediction head that directly maps temporal embeddings to predictions.
Our default configuration includes:
\begin{itemize}
\item Embedding layer: An MLP layer mapping meta features to 1024-dimensional latent embeddings, followed by a temporal encoding layer similar to that in \cite{zhou_informer_2021}.
\item Temporal encoder: An MLP-Mixer \cite{tolstikhin_mlp-mixer_2021} alternately applying to the last two dimensions.
\item Cross-sectional encoder: A multi-head self-attention \cite{vaswani_attention_2017} layer with increased dropout probability to promote sparsity of connections among stocks.
\item Position sizer: An LSTM operating on the first dimension to encode sequential information across multiple holding periods.
\end{itemize}

\tableWhatToAdd

\paragraph{Evaluation}
We employed several portfolio metrics to assess performance:
\begin{itemize}
\item \textbf{Annualized return} and \textbf{maximum drawdown}: Measuring profitability and risk-control capability, respectively.
\item \textbf{Sharpe ratio} and \textbf{Calmar ratio}: Characterizing risk-adjusted profitability.
\end{itemize}
All reported metrics are averaged over four repeated runs and computed as excess values relative to the CSI1000 benchmark. We conducted backtests using Qlib, incorporating a 0.3\% transaction cost. Trading was simulated at market close, with stocks hitting the daily change limit marked as non-tradable.

For a more comprehensive description of the experimental setup, please refer to the appendix.

\subsection{Results}
\subsubsection{Overall Effectiveness}
We evaluated the performance of various portfolio construction approaches, comparing traditional multi-stage methods with guided learning techniques. For multi-stage approaches, we considered two settings: "Optimization," which applies a portfolio optimizer to model predictions, and "NN," which appends a neural network to trained predictive models. In the realm of end-to-end modeling approaches, we explored four variants. The "Guide-free" setting serves as our baseline, employing no additional guidance. The "Multi-task" setting incorporates the Information Coefficient (IC) as a guide added to the cross-sectional embedding. The "IC-Guide" approach adds the IC guide to the temporal embedding. Lastly, the "IC-Guide + Return Guide" setting combines the IC guide with an additional cross-sectional return guide on the cross-sectional embedding.

Results in Table \ref{tab:overall_effectiveness} demonstrate that guide-free end-to-end learning outperforms traditional stage-wise methods, achieving higher annualized returns (23.19\% vs 17.97\% for optimization-based and 12.41\% for NN-based) and better risk-adjusted metrics (Sharpe ratio of 0.94 vs 0.73 and 0.52 respectively).
The introduction of IC-Guide further enhances performance, particularly in controlling maximum drawdown (-7.70\% vs -8.90\% for guide-free) and improving the Calmar ratio (3.27 vs 2.58). Notably, the multi-task approach and the addition of a Return Guide to IC-Guide did not yield further improvements, suggesting that the benefits of guided learning are sensitive to the specific implementation. The stage-wise NN setting showed decreased performance compared to the optimization approach, possibly due to conflicting training objectives.

\subsubsection{Generalization}
We examined the generalizability of guided learning across different model architectures.
Our experiment involved three representative architectures as temporal encoders: LSTM for RNN, TCN for CNN, and PatchTST for Transformers. We applied the same IC-Guide to the temporal embedding of each model.
Table \ref{tab:model_generalize} presents the results, demonstrating the effectiveness of the IC-Guide approach across various architectures. LSTM models showed improvements in all metrics, with a notable 12.30\% increase in Calmar ratio. TCN models exhibited enhanced performance, particularly in Sharpe ratio (6.59\% improvement) and Calmar ratio (11.84\% improvement). PatchTST models showed modest gains in annualized return and Sharpe ratio, potentially requiring more tuning of the model configuration. The consistent performance enhancement across diverse models indicates that guided learning generalizes well across different model architectures.

\subsubsection{Influence of Guide Placement}
Table \ref{tab:where_to_add} illustrates the effect of applying the IC-Guide at different stages of the model. Temporal embedding guidance yields the best overall performance, with the highest Sharpe ratio (0.88) and Calmar ratio (2.73). This suggests that guiding the model at the temporal level allows it to better capture time-dependent patterns crucial for portfolio optimization.
Embedding-level guidance performs well, particularly in annualized return (20.75\%), indicating its effectiveness in enhancing the model's ability to extract meaningful features from raw data. However, its slightly lower Sharpe and Calmar ratios compared to temporal guidance suggest that it may not optimize the risk-return trade-off as effectively.
Cross-sectional guidance shows the least improvement, with lower performance across all metrics. This could be because later-stage guidance may constrain the model too much, limiting its ability to learn complex cross-asset relationships beyond accurate IC prediction. These results suggest the importance of guide placement.

\subsubsection{Guide Type Comparison}
Table \ref{tab:what_to_add} compares different types of guides. Mean Squared Error (MSE) guidance yields the best overall performance, with the highest annualized return (18.94\%) and Calmar ratio (2.06). Classification (CLF) guidance shows comparable performance to MSE, with slightly lower metrics across the board. Ranking-based guidance significantly underperforms, suggesting that this approach may not align well with the portfolio optimization objective. These results highlight the importance of choosing appropriate guidance objectives that align with the end goal.

\subsubsection{Parameter Sensitivity}
We also studied the parameter sensitivity of guided learning to verify its effectiveness in lubricating end-to-end modeling. See more experimental results in the appendix. 

\section{Related Work}

\paragraph{End-to-end Modeling}
With the growth in data and computational resources, end-to-end modeling has become increasingly prevalent in many domains, superseding multi-stage approaches that follow the predict-then-optimize paradigm \cite{elmachtoub_smart_2022}. For instance, in autonomous driving, \cite{hu_planning-oriented_2023} proposes a Transformer-based end-to-end model that generates planned trajectories directly from perceptual input. This model is trained using both final planning utility and intermediate vision-based losses, including occlusion prediction and object detection. In quantitative investment, \cite{wei_e2eai_2023} introduces an end-to-end deep learning framework that generates positions from raw stock data, employing a training approach that combines final portfolio metrics with intermediate losses incorporating feature selection and inter-stock relation modeling. Other works \cite{liu_deep_2023, nagy_generative_2023} also explore end-to-end approaches, demonstrating the growing trend towards unified modeling strategies.

\paragraph{Manipulating Intermediate Representations}
The concept of incorporating intermediate supervision in neural network training has a rich history. Deep supervision \cite{lee_deeply-supervised_2015}, extensively studied in computer vision \cite{shen_object_2019, ren_deepmim_2023, zhang_contrastive_2022}, aims to stabilize neural network training by applying auxiliary (self-)supervised losses to intermediate embeddings. Recently, similar principles of interpreting and controlling intermediate representations have been explored in large language models \cite{gao_scaling_2024, templeton2024scaling}. These studies utilize sparse autoencoders to map intermediate representations of large-scale Transformer models to discrete "concept" vectors. Notably, \cite{templeton2024scaling} demonstrated that manipulating these intermediate representations at the concept level can lead to controlled outputs, offering potential benefits in scenarios such as AI safety.
\section{Conclusions and Future Works}
To conclude, guided learning has the potential to enhance training stability, performance, and interpretability across various complex real-world AI applications. Meanwhile, we currently envision several key aspects for further investigation: 
\begin{enumerate}
    \item \textbf{Designing guides:} Current approaches rely on ad-hoc manual design with domain expertise. Future research may focus on developing principled methods for creating effective guides, including automated techniques and investigating their transferability across different applications.
    \item \textbf{Analyzing guide effectiveness:} A deeper understanding of why guided learning works is crucial. This includes theoretical analyses of how guides influence the optimization landscape and empirical studies on their impact on model convergence and performance. Such investigations could potentially draw from optimization theory and information geometry to provide a solid theoretical foundation for guided learning. Exploring the interplay between guides and the primary optimization goal could lead to more nuanced strategies, potentially uncovering synergies that enhance overall system performance. Furthermore, comparative studies across various problem domains could help identify the characteristics of effective guides in different contexts.
    \item \textbf{Integration with more scenarios:} Potential applicability to other complex, end-to-end scenarios such as autonomous driving, and robotics control. These domains share characteristics that make them suitable for guided learning, which could offer a middle ground between traditional pipelines and end-to-end paradigms. 
\end{enumerate}

\bibstyle{aaai25}
\bibliography{references}

\newpage
\appendix

\figurePsen
\section{Additional Experimental Details}
\subsection{Dataset}
We used historical dataset of Chinese A-share market, including over 4200 stocks from 2013-01-01 to 2022-01-04. Features in this dataset contains volume-price data, fundamental data, and dummy variables such as industry and country. We set the lookback window to 10 days and predict horizon as 1 day, meaning that at each day after market close, we take the 10-day historical features of each stock and predict their 1-day forward close-to-close return tomorrow.

\paragraph{Data preprocessing}
All features are first wonorized with the upper and lower limit set to 0.1 times the cross-sectional sample median, in order to remove outliers. Then the data is transformed into cross-sectional z-scores with a clip bound of 3. Finally, the NaNs in input features are filled with 0s. 

\paragraph{Sampling}
We filtered out samples that are non-tradable in each trading day.
For multi-stage decision models, we applied the cross-sectional sub-sampling technique. At each iteration, a trading day is randomly sampled. Then we randomly pick 80\% of all the stocks on that trading day as a cross-section sample. Compared with full cross-sectional sampling, such sampling greatly enriches the sample size. For guided learning models, we picked 22 consecutive trading days to compute the Sharpe ratio throughout the whole period, and the sub-sampling ratio is adjusted to 10\% due to GPU memory limit.

\subsection{Model Details}
We trained the model with learning rate of 1e-4 and a maximum epoch number of 200, with early-stopping patience of 10. Models are validated each 0.5 training epochs. The temporal encoder has 3 layers. The cross-sectional has only 1 layer, and the LSTM model also has 1 layer. All model parameter counts total up to 8 million.

\subsection{Evaluation Details}
Let $r_t$ denote the excess return of a portfolio over the benchmark at day $t \in \{1, ..., T\}$. Then the annualized return is computed as
\begin{equation*}
    \text{Annualized Return} = \frac{238}{T} \sum_{t=1}^T r_t
\end{equation*}
where 238 indicates the number of trading days in a year. The maximum drawdown is computed as
\begin{align*}
    & \text{Maximum Drawdown} \\
    &= \min_{t \in \{1, \dots, T\}} \left( \sum_{i=1}^t r_i - \max_{j \in \{1,\dots,t\}} \sum_{i=1}^j r_i \right)
\end{align*}
The mean and standard deviation of daily excess returns is computed as
\begin{equation*}
    \mu =  \frac{1}{T} \sum_{t=1}^T r_t
\end{equation*}
\begin{equation*}
    \sigma = \sqrt{\frac{1}{T-1} \sum_{t=1}^T \left( r_t - \mu \right)^2}
\end{equation*}
where \( \bar{r} \) is the average daily excess return. The Sharpe ratio is then computed as
\begin{equation*}
    \text{Sharpe Ratio} = \sqrt{238} \cdot \frac{\mu}{\sigma}
\end{equation*}
Finally, the Calmar ratio is computed as
\begin{equation*}
    \text{Calmar Ratio} = \frac{\text{Annualized Return}}{\text{Maximum Drawdown}}
\end{equation*}
The Calmar ratio measures the risk-adjusted return of a portfolio by comparing the annualized return to the maximum drawdown. A higher Calmar ratio indicates a better performance considering the drawdown risk.

\subsection{Implementation details}
All experiments are carried out on a computer with two 64-core CPUs, 2TB main memory, and one NVIDIA A100 GPU with 40GB HBM memory. All models are implemented using PyTorch.

\section{Addtional Results}
\subsection{Parameter Sensitivity}
Figure \ref{fig:psen} illustrates the parameter sensitivity of four key performance metrics (annualized return, maximum drawdown, Sharpe ratio, and Calmar ratio) to guide coefficients in the IC Guide + Return Guide setting. All metrics exhibit complex, non-monotonic relationships with both the Return Coefficient and IC Coefficient, highlighting the challenges of parameter tuning. Generally, optimal performance across metrics tends to occur when both coefficients are in the mid-range (approximately 1.4-2.0), suggesting a balanced approach often yields the best results.

\end{document}